\documentclass[letterpaper, 10 pt, conference]{ieeeconf}  

\IEEEoverridecommandlockouts                              
\usepackage[T1]{fontenc}

\usepackage{hyperref}
\hypersetup{
    colorlinks=true,
    linkcolor=blue,
    filecolor=magenta,      
    urlcolor=cyan,
    pdftitle={Overleaf Example},
    pdfpagemode=FullScreen,
    }

\usepackage{cite}
\usepackage{amsthm}
\usepackage{amsmath} 
\usepackage{amssymb}  
\usepackage{amsfonts}
\usepackage{mathtools}

\usepackage{graphicx}
\usepackage[font=small]{caption}
\usepackage{subcaption}
\usepackage{capt-of}  
\usepackage{cuted} 
\usepackage{textcomp}
\usepackage{xcolor}
\usepackage{stmaryrd}
\usepackage[scr=boondox]{mathalfa}
\usepackage[flushleft]{threeparttable}
\usepackage{array}
\usepackage{makecell}
\usepackage{multirow}

\usepackage{algpseudocode}
\usepackage[ruled,lined]{algorithm2e}

\usepackage{todonotes}

\theoremstyle{remark}
\theoremstyle{definition}

\newtheorem{definition}{Definition}[section]

\DeclareMathOperator*{\Min}{min}

\usepackage[T1]{fontenc}
\usepackage[utf8]{inputenc}
\usepackage{tabularx,ragged2e,booktabs, caption}
\newcolumntype{C}[1]{>{\Centering}m{#1}}

\usepackage{setspace}
\usepackage{todonotes}
\usepackage{array}

\newcolumntype{R}{@{\extracolsep{3cm}}r@{\extracolsep{0pt}}}%

\mathtoolsset{showonlyrefs}
\DeclareMathAlphabet{\mathpzc}{OT1}{pzc}{m}{it}

\newcommand{\rd}{l_d} 
\newcommand{\rl}{l}
\newcommand{\rt}{r_t}
\newcommand{\li}{d^i}
\newcommand{\lo}{d^1}
\newcommand{\lm}{d^n}

\makeatother
\makeatletter
\title{\LARGE \bf Efficient Domain Coverage for Vehicles with Second-Order Dynamics via Multi-Agent Reinforcement Learning}

\author{Xinyu Zhao, Razvan C. Fetecau, Mo Chen
\thanks{X. Zhao and R. C. Fetecau are with the Department of Mathematics, Simon Fraser University, Burnaby, BC, Canada (SFU) {\tt\small \{xza261@, van@math.\}sfu.ca}). M. Chen is with the School of Computing Science, SFU, {\tt\small mochen@cs.sfu.ca}.}%
\thanks{This work received support from Huawei Technologies Co., Ltd.}%
}

\begin{document}
\maketitle
\thispagestyle{empty}
\pagestyle{empty}

\begin{abstract}
Collaborative autonomous multi-agent systems covering a specified area have many potential applications. 
Traditional approaches for such problems involve designing model-based control policies; however, state-of-the-art classical control policy still exhibits a large degree of sub-optimality.
We present a combined reinforcement learning (RL) and control approach for the multi-agent coverage problem involving agents with second-order dynamics, with the RL component being based on the Multi-Agent Proximal Policy Optimization Algorithm (MAPPO).
Our proposed network architecture includes the incorporation of LSTM and self-attention, which allows the trained policy to adapt to a variable number of agents.
Our trained policy significantly outperforms the state-of-the-art classical control policy.
We demonstrate our proposed method in a variety of simulated experiments.
\end{abstract}

\section{INTRODUCTION}

Multi-agent cooperative area coverage is a promising research field, with the objective being to deploy a set of agents over a domain of interest to achieve optimal sensing. Applications include monitoring \cite{1, 2}, target detection \cite{3, 4}, and search and rescue \cite{5, 6}. In the past decades, researchers have proposed various approaches, including classical mathematical model-based and learning-based methods, to solve coverage problems. 

Classical control presents many sophisticated control laws for coverage problems under certain assumptions. Common approaches include solving an optimization problem that involves Voronoi tesselations\cite{lo1, lo2}, potential field methods\cite{pfm1, CHACON202015167}, and scalar field mapping\cite{sfm2, sfm1}. Although those classical approaches have proven effective in many applications, designing a suitable model under complex scenarios is always challenging. Thus, the learning-based method has drawn increasing attention in recent years.

\begin{figure}[ht]
\centerline{\includegraphics[width=0.45\textwidth]{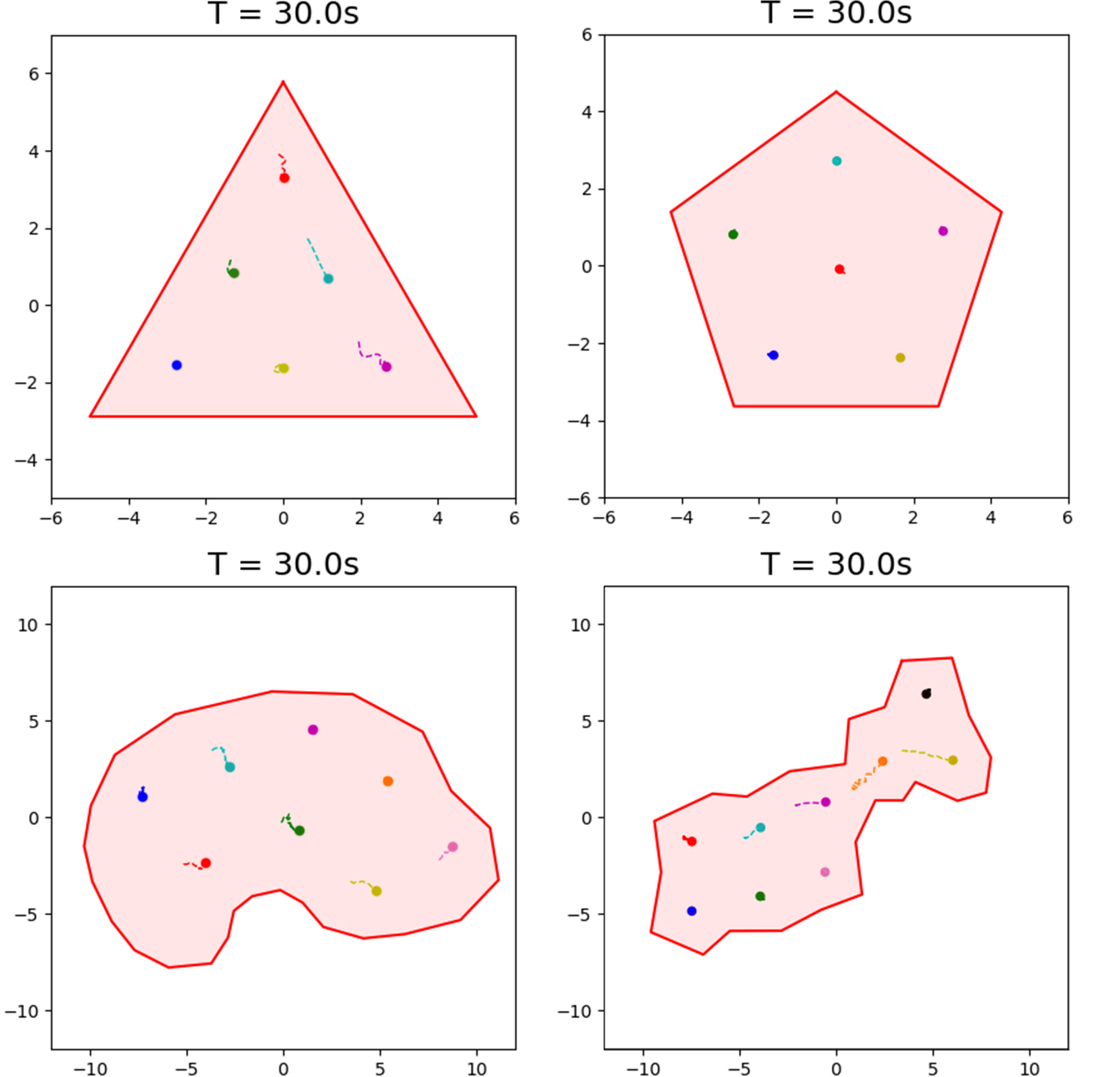}}
\caption{The final coverage configuration of our policy for 6, 8, and 9 agents in both convex and non-convex polygon domains. The simulation duration is the 30s and the tails in all plots represent the agents' trajectories in the last 15s of the simulation. Small perturbations over the last 15s indicate reliable persistent coverage performance.}
\label{fig:DifferentDomain}
\end{figure}

For the learning-based approach,  the recent deep reinforcement learning (RL) algorithm could leverage the advantages of neural networks (NNs) and adaptive learning capabilities to obtain a policy that optimizes the performance metric over the trajectory.
NNs have the ability to approximate a broad class of functions, and through RL, a control policy parameterized by a NN can be directly learned through real-time interactive feedback from the environment, hence avoiding the challenge of choosing an accurate model.

Many recent works have investigated the RL method for coverage in discrete space\cite{MARL_COV_2, MARL_COV_3, MARL_COV_4}, but works on applying RL to coverage problems with continuous state and action spaces are still limited. One related work \cite{Adepegba2016MultiAgentAC} applies an actor-critic method involving a specific value function formulation to multi-agent continuous coverage control. In \cite{Kouzehgar_2020}, the authors used MADDPG \cite{MADDPG} to learn a coverage policy for the ocean monitoring task. Another limitation in previous work is that the NN policy can only handle a fixed number of agents due to the typical multiple-layer perceptron (MLP) networks used in this domain, which require a fixed-dimension input. In addition, besides achieving coverage, another performance metric, such as the time efficiency of achieving coverage, has not been considered.

We propose a novel RL method to address some of the limitations of the previous work. Our target is to learn a policy for continuous state-action space, which 1) allows the agents to achieve the final coverage configuration in a time-efficient manner and 2) generalizes to scenarios with a variety of domains and a varying number of agents.
For illustration, Fig. \ref{fig:DifferentDomain} shows the final coverage configurations achieved with our policy, across different numbers of agents, for regular polygon and non-convex polygon domains. To our best knowledge, this is the first work on adaptive, efficient coverage control using RL for variable numbers of agents. Our contributions are as follows:
\begin{itemize}
    \item We propose an RL approach for continuous control of area coverage with second-order dynamics through potential-based reward shaping.
    \item We propose a new training scheme involving an LSTM-based policy and self-attention-based value network to allow the agent to generalize to scenarios involving varying numbers of agents.  
    \item Our approach results in successful and more time-efficient coverage compared to prior work.    
\end{itemize}
The rest of the paper is organized as follows. Section \ref{sect:problem} states the problem formulation for domain coverage. Section \ref{sect:background} provides background information. Section \ref{sect:approach} details our method for solving the time-efficient coverage control across different numbers of agents. Section \ref{sect:experiments} illustrates our approach with simulations.

\section{Problem Formulation}
\label{sect:problem}

We consider a set of $n$ homogeneous agents indexed by $I = \{1, \dots, n\}$, with dynamics given by
\begin{equation}
\label{eqn:dynamics}
\begin{cases}
\dot{p}^i = v^i, \qquad \lVert v^i \rVert \leq v_{max},\\
\dot{v}^i = a^i, \qquad \lVert a^i \rVert \leq a_{max},\\
\end{cases}
\end{equation}

\noindent where $p^i = (p_x^i, p_y^i)$, $v^i = (v_x^i, v_y^i)$ and $a^i = (a_x^i, a_y^i)$ are the position, velocity and the control input of the $i$th agent, respectively. Also, $v_{max}$ and $a_{max}$ are bounds on velocities and control inputs, and $\|\cdot\|$ denotes the Euclidean distance.

The problem definition of multi-agent area coverage can vary under different scenarios and assumptions. 
In this paper, we are interested in deploying the set of agents into a target domain to achieve uniform area coverage.
Specifically, we adopt the concept of coverage presented in \cite{CHACON202015167,https://doi.org/10.48550/arxiv.2009.12211}.
\begin{definition}
\label{defn:rd-cover}
We say that the group of agents $I$ is an $\rl$-\textit{subcover} configuration for a compact domain $\Omega\subset \mathbb{R}^2$ if
\begin{enumerate}
    \item[1.] $B_{\frac{\rl}{2}}(i) \subset \Omega $ for all $i \in I$, \vspace{0.15cm}
    \item[2.] $\|p^i - p^j\| \geq \rl$ for all pairs $i, j \in I$, $i \neq j$.
\end{enumerate}

\noindent where $B_{\frac{\rl}{2}}(i)$ denotes the ball of radius $\frac{\rl}{2}$ centered at $p^i$.
\end{definition}

The main interest in \cite{CHACON202015167} is in $\rd$-\textit{subcover} configurations, where $\rd = \sqrt{\frac{Area(\Omega)}{n}}$,
based on the assumption that each agent covers the same amount of square area. Aside from reaching an $\rd$-\textit{subcover} configuration, we are also interested in the time efficiency of achieving domain coverage. Based on Def. \ref{defn:rd-cover} and the time efficiency concerns, the primary interest of this work is defined as follows.
\medskip

\noindent
\textbf{Multi-Agent Efficient Domain Coverage Control:} We aim to find a control policy that drives a set of $n$ agents from any initial positions to an $\rd$-\textit{subcover} of some compact domain $\Omega$ as quickly as possible.

\section{Background}
\label{sect:background}
In this section, we review some important background information for this work. We consider multi-agent domain coverage control as a fully cooperative task with homogeneous agents, formulated as a decentralized partially observable Markov decision process (Dec-POMDP) \cite{10.5555/2967142} of $n$ agents consisting of a tuple of $\langle S, A, O, P, R, n, \gamma \rangle$. 
Here, $S$ is the joint state space of all agents, 
$A$ is the joint action space, 
$O$ is the joint observation space, 
$P = \mathbb{P}(s_{t+1}| s_t, a)$ is the transition probability function from current joint state $s_t$ to the next state $s_{t+1}$ given the joint action $a=(a^1,\dots,a^n)$, 
$R(s_t, a, s_{t+1}) \colon S \times A \times S \longrightarrow \mathbb{R}$ is the global reward function, 
$n$ is the number of agents, and 
$\gamma\in [0, 1)$ is a constant discount factor.

The joint state $s$ is only partially observable to each agent, who draws the local observation $o^i$ according to its own observation function $\eta^i \colon S \rightarrow O$. 
At time $t$, each agent samples an action $a^i_t$ from a stochastic policy $\pi^i(a^i_t|o^i_t)$ given $o^i_t$. 
By executing the joint action $a_t = (a^1_t, \dots, a^n_t)$, the joint state evolves from $s_t$ to $s_{t+1}$ according to the transition probability function $P$, and an immediate reward $R_{t} = R(s_t, a_t, s_{t+1})$ is given by the environment. The return $G_t$ is defined as the discounted cumulative future reward starting from state $s_t$, given by $G_t = \sum_{k=0}^{T} \gamma^k R_{t+k}$. 
The objective for all agents is to find a joint policy $\pi = (\pi^1,\dots, \pi^n)$ that maximizes the corresponding state value function $V_{\pi}(s) = \mathbb{E}_{\pi}\big[ G_t | s_t = s \big]$ and action value function $Q_{\pi}(s, a) = \mathbb{E}_{\pi}\big[ G_t | s_t = s, a_t = a \big]$ for any given state $s$ and action $a$.
\begin{figure*}
    \centering
    \vspace*{0.175cm}
    \includegraphics[width=0.9\textwidth]{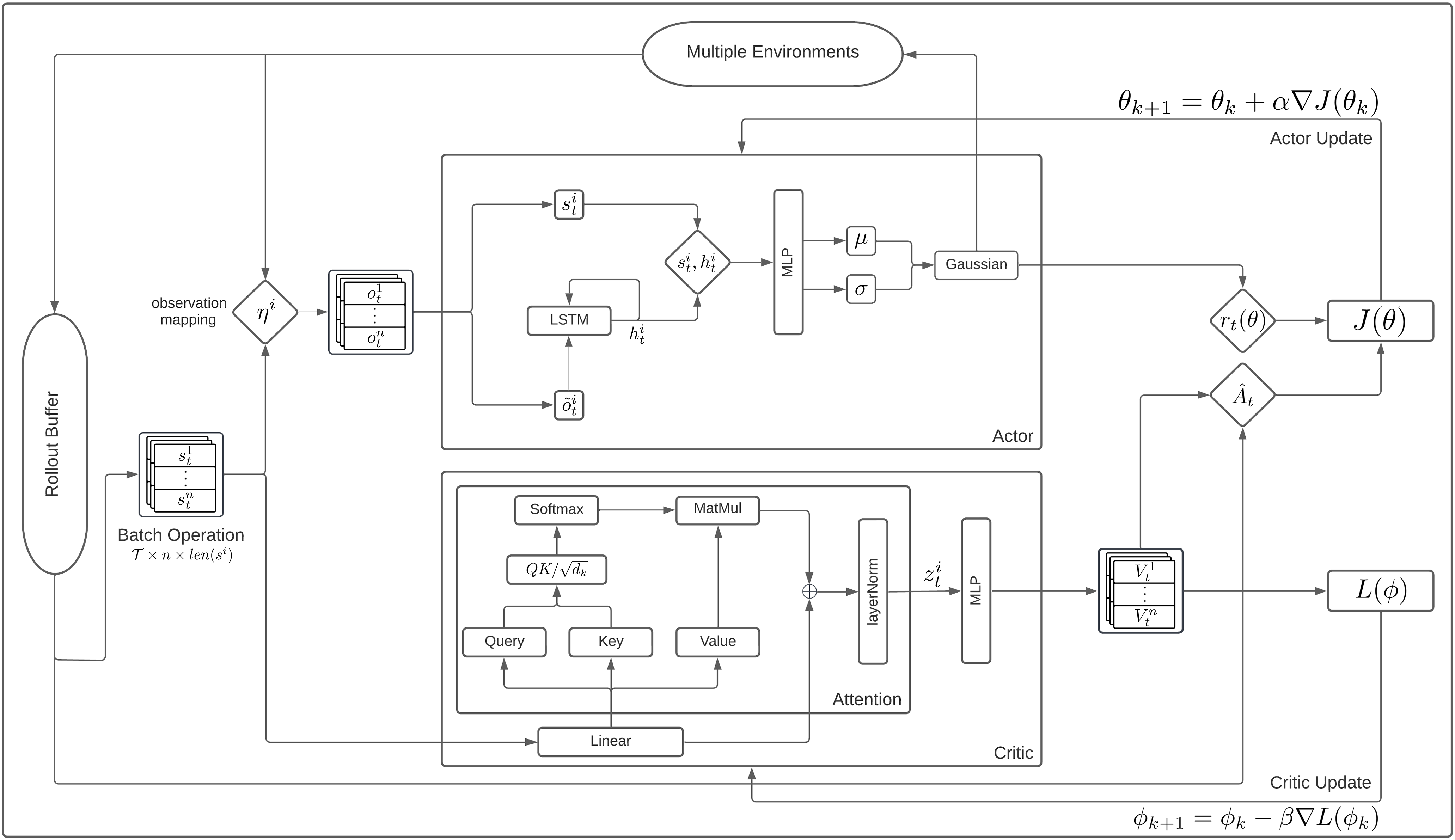}
    \caption{An overview of the architecture of our approach.}
    \label{fig:img1}
\end{figure*}
\allowdisplaybreaks

\subsection{Multi-Agent Proximal Policy Optimization}
\label{subsect:MAPPO}
Our approach is built upon multi-agent proximal policy optimization (MAPPO) \cite{MAPPO}, an actor-critic algorithm for the MARL problem. The main feature of MAPPO is to extend the PPO algorithm \cite{PPO} under the centralized training and decentralized execution (CTDE) framework \cite{MADDPG}. In particular, the CTDE extension for any actor-critic architecture allows the critic network to access the environment's global state, while the policy network for every agent still chooses the action based on the agent's local observation only.

Consider an $n$-agent environment described by Dec-POMDP, with a set of policies $\pi_{\theta} = (\pi_{\theta^1}, \dots, \pi_{\theta^n})$ parametrized by $\theta =  (\theta^1, \ldots, \theta^n)$, and a centralized value function $V_{\phi}$ parametrized by $\phi$. The MAPPO algorithm samples trajectories from an old joint policy $\pi_{\theta_{old}}$, and updates the policy parameter for each agent by maximizing the following surrogate objective \begin{equation}
\label{eqn:max-E}
\begin{aligned}[b]
J(\theta^i) &= \mathbb{E}_{s \sim \rho_{\pi_{\theta_{old}}}, a^i \sim \pi_{\theta_{old}^i}} \Bigl[\Min \bigl(\rt(\theta^i)A_{\pi_{\theta_{old}}}(s_t, a_t),\\
& \quad\quad\quad\quad\;\; \text{clip}(\rt(\theta^i), 1+\epsilon, 1-\epsilon)A_{\pi_{\theta_{old}}}(s_t, a_t) \bigr) \Bigr],
\end{aligned}
\end{equation}where $\rho_{\pi_{\theta_{old}}}$ is the state visitation frequency induced by old policy $\pi_{\theta_{old}}$, $\rt(\theta^i)$ is the probability ratio calculated as
\begin{equation}
\label{eqn:ratio}
\rt(\theta^i) = \frac{\pi_{\theta^i}\big(a_t^i|\eta^{i}(s_t)\big)}{\pi_{\theta_{old}^i}\big(a_t^{i}|\eta^{i}(s_t)\big)},   
\end{equation}$A_{\pi_{\theta_{old}}}(s_t, a_t)$ is the advantage value at time $t$, defined as $A_{\pi_{\theta_{old}}}(s_t, a_t) = Q_{\pi_{\theta_{old}}}(s_t, a_t) - V_{\pi_{\theta_{old}}}(s_t)$ and $\epsilon \in [0, 1)$ is a hyperparameter controlling the clip rate. 

The advantage function $A_{\pi_{old}}(s_t, a_t)$ can be estimated in several ways as in \cite{GAE} based on the centralized value function $V_{\phi}$. Also, $V_{\phi}$ is found by minimizing the standard mean squared error loss (MSE) between $V_{\phi}(s_t)$ and a value target $\hat{V}_t$ such as a Monte-Carlo estimation or temporal-difference target. To keep the notation simple, the advantage $A_{\pi_{old}}(s_t, a_t)$ is simplified as $\hat{A}_t$ in the rest of paper. For more details about the PPO and MAPPO objective, we refer the reader to \cite{PPO, MAPPO}. 

\subsection{Behaviour Cloning}
\label{subsect:BC}

The behavior cloning (BC) technique is also used to help the training in this work. The main idea of BC is to copy an expert's policy using supervised learning. For a stochastic policy $\pi_{\theta}$, BC minimizes the error between the expert's action and the maximum likelihood action over the current policy. This can be done via several different objective functions. Our approach uses a simple version of BC, where the objective is given by 
\begin{equation}
\label{eqn:bc_loss}
L^{BC}(\pi_{\theta}) = \frac{1}{T}\sum_{t=0}^{T} \bigr\|\hat{a}^i_t - \mu_{\pi_{\theta}}(o^i_t)\bigr\|^2 ,
\end{equation}
\noindent where $\hat{a}^i_t$ is the expert's action under the observation $o^i_t$, and $\mu_{\pi_{\theta}}(o^i_t)$ is the mean of stochastic policy $\pi_{\theta}$.

\section{Approach}
\label{sect:approach}

In this section, we explain the details of our approach for applying MARL to solve the multi-agent efficient domain coverage control problem defined in Sec. \ref{sect:problem}. 
Our approach addresses two major challenges: The learned policies need to 1) produce a time-efficient trajectory while completing the coverage task, and 2) be able to control different numbers of agents in the environment. 

To address the two challenges above, we build our method based on the MAPPO algorithm with newly designed policy and value network architectures. In order to guide the agent to learn a suitable control policy that solves the multi-agent coverage problem efficiently, a proper reward function needs to be designed. We accomplish this by choosing a suitable potential function from the classical coverage control. With an adequately shaped reward, agents learning to maximize the return also improve time efficiency.  

To enable adaptive control of our policy over variable numbers of agents, we take the idea from natural language processing to reshape the structure of both actor and critic network by leveraging the long short-term memory (LSTM) \cite{10.1162/neco.1997.9.8.1735} and the self-attention \cite{https://doi.org/10.48550/arxiv.1706.03762} mechanism to encode the state information into an ideal format as input for MLP networks. Meanwhile, the value decomposition (VD) \cite{VDN} is applied as a natural complement to the self-attention structures to facilitate the training further. The new architecture enables the training process to happen concurrently across multiple environments with a variable number of agents. 
An overview of our approach is illustrated in Fig. \ref{fig:img1}. 


\subsection{Action Space, Agent Observation and Environment State}
\label{subsect:states}
In this work, we aim to directly output the continuous control force following the system \eqref{eqn:dynamics}. Namely, the action space in this work is defined as the agent's acceleration. The assumptions on the environment state and agent observation in this work are as follows. We assume that every agent knows its position $p^i$ and velocity $v^i$. We further assume that every agent can measure its position relative to the domain boundary $\partial \Omega$, given by the vector
\begin{equation}
\label{eqn:hi}
\li: = p^i - P_{\partial \Omega}(p^i),
\end{equation}
where $P_{\partial \Omega}$ denotes the projection operator to $\partial \Omega$. With these assumptions, we define the agent's internal state $s^i$ to be
\begin{equation}
s^i = \bigr(p^i, \, v^i, \, \hat{\li}, \, \|\li\|, \, \mathbb I (p^i)\bigr),   
\end{equation}
where $\hat{\li}= \li/\|\li\|$ and $\mathbb I(p^i)$ is given by
\begin{equation}
\mathbb I (p^i)=
\begin{cases}
\hspace{0.25cm} 1, & \text{ if } p^i \not\in \Omega, \\
-1, & \text{ if } p^i \in \Omega.
\end{cases}
\end{equation}
We further construct the environment full state $s$ by stacking all $s^i$ as
\begin{equation}
s = 
\begin{bmatrix}
s^1\\
\vdots\\
s^n
\end{bmatrix}=
\begin{bmatrix}
p^1 & v^1 & \hat{\lo} & \|\lo\| & \mathbb I (p^1)\\
\vdots & \vdots & \vdots & \vdots & \vdots\\
p^n & v^n & \hat{\lm} & \|\lm\| & \mathbb I (p^n)
\end{bmatrix}.
\end{equation}
Expressing the state as a matrix will be useful in Sec. \ref{subsect:Critic}.

Partial observability in this paper means that agents do not know all the information about the other agents. However, we assume that each agent knows its position relative to the other agents, represented by $p^{i,j} = p^i - p^j$ for $i\neq j$. Therefore, we consider the agent local observation $o^i = (s^i, \tilde{o}^i)$ to consist of two parts, where the first part is the agent’s internal state $s^i$ and the second part is the observation of other agents relative to itself, given by
\[
\tilde{o}^i = \bigr( p^{i,1},p^{i,2},\dots,p^{i,i-1},p^{i,i+1},\dots,p^{i,n} \bigr).
\]
In practice, such an observation can be obtained by ranging sensors such as LIDAR.


\subsection{Reward Shaping}
\label{subsect:reward}
We now discuss reward shaping for the domain coverage problem stated in Sec. \ref{sect:problem}. We design the reward function to address the coverage goal, as well as time efficiency. We base our reward function on the artificial potentials used in \cite{CHACON202015167,https://doi.org/10.48550/arxiv.2009.12211} for the dynamics of \eqref{eqn:dynamics}.
Specifically, the potential functions in \cite{CHACON202015167,https://doi.org/10.48550/arxiv.2009.12211} for agent-domain and inter-agent interactions, are given respectively by \eqref{eqn:Vh} and \eqref{eqn:VI}:
\begin{equation}
\label{eqn:Vh}
U_h(p^i) = 
\begin{cases}
0,   \text{ for } \llbracket \li \rrbracket \leq -\frac{\rd}{2}, \\[0.2pt]
\frac{1}{2}(\llbracket \li \rrbracket + \frac{\rd}{2})^2,  \text{ for } \llbracket \li \rrbracket > -\frac{\rd}{2},
\end{cases} 
\end{equation}
\begin{equation}
\label{eqn:VI}
U_{I}(p^{i,j}) =
\begin{cases}
\frac{1}{2}(\|p^{i,j}\| - \rd)^2,  \text{ for } \lVert p^{i,j} \rVert < \rd, \\[0.2pt]
0,   \text{ for } \lVert p^{i,j} \rVert \geq \rd.
\end{cases} 
\end{equation}
Here, $\llbracket \li \rrbracket$ denotes the signed distance of $p^i$ from $\partial \Omega$, i.e.,
\begin{equation}
\llbracket \li \rrbracket = \mathbb{I}(p^i) \cdot \|p^i - P_{\partial \Omega}(p^i)\|.
\end{equation}

Note that the expressions of $U_h$ and $U_I$ correspond directly to the two conditions in Def. \ref{defn:rd-cover} (for $\rl=\rd$). 

The total potential energy of the system is given by
\begin{equation}
\label{eqn:energy-total}
\Phi = \sum_{i}^{n} \Phi^i,
\end{equation}
\noindent where $\Phi^i$ is the individual potential obtained by summing up $U_h$ and $U_I$ as
\begin{equation}
\label{eqn:energy-i}
\Phi^i = 2U_{h}\left(p^{i}\right) + {\sum_{j\neq i}^{n}}U_{I}\left(p^{i,j}\right).
\end{equation} 

The global minimum of $\Phi$ is zero and attained at $\rd$-\textit{subcover} configurations, and is a measure for coverage performance.  
In our method, we shape the reward by transforming the potential energy slightly. We define the agent's individual reward $R^{i}(s_{t}, a_{t}, s_{t+1})$ as 
\begin{equation}
\label{eqn:reward-i}
R^i(s_{t}, a_{t}, s_{t+1}) =
\begin{cases}
- M, & \text{if agent $i$ is outside } \Omega,\\[2.5pt]
- \Phi^i_{t+1}, & \text{if agent $i$ is inside } \Omega.
\end{cases}
\end{equation}

\noindent where $M$ is a positive constant, and $\Phi_{t+1}^{i}$ denotes the individual potential for agent indexed by $i$ at time step $t+1$. Here, a constant reward $-M$ is assigned to an agent that is outside the domain, as the potential could be arbitrarily large in such case. We set the constant $M$ to be
\[
M = \sup\{\Phi:\text{all agents are inside } \Omega\}.
\]

We further define $R$ as the global reward for the Dec-POMDP setup, which is calculated by taking the sum of all agents' individual rewards, i.e.,
\begin{equation}
\label{eqn:reward-additive}
R(s_t, a_t, s_{t+1}) = \sum_{i=1}^{n} R^i(s_{t}, a_{t}, s_{t+1}).
\end{equation}

Let $R^i_t = R^i(s_{t}, a_{t}, s_{t+1})$, and $R_t = R(s_t, a_t, s_{t+1})$. The formulation in \eqref{eqn:reward-additive} allows each agent to receive a different reward, while simultaneously knowing its contribution to the team. We remark that the additive reward formulation allows us to use direct VD, which facilitates the training process in our approach. Details on using the reward $\eqref{eqn:reward-additive}$ will be presented in Sec. \ref{subsect:Critic}.

The reward in \eqref{eqn:reward-additive} also naturally encourages time efficiency. Indeed, the maximum of $R$ is zero and can be attained at any $\rd$-\textit{subcover} configuration; hence, a trajectory that attains the coverage goal faster will gain higher returns. 


\subsection{Actor Architecture for Variable Length Input}
\label{subsect:Actor}

A typical policy network architecture in MARL uses the MLP networks for each agent. However, such architecture has difficulties to adapt into the scenario of a variable number of agents. In this work, we utilize the parameter sharing (PS) technique \cite{10.1007/978-3-319-71682-4_5, https://doi.org/10.48550/arxiv.2005.13625}, where all homogeneous agents share a single policy network $\pi_{\theta^1} = \dots = \pi_{\theta^n}$, resulting in the policy objective being reduced to follows:
\begin{equation}
\label{eqn:clip}
J(\theta)  = \mathbb{E}_{\pi_{\theta}}\bigr[\Min (\rt(\theta)\hat{A}_t, \text{clip}(\rt(\theta), 1-\epsilon, 1+\epsilon)\hat{A}_t) \bigr],
\end{equation}and the probability ratio $\rt(\theta)$ preserving same definition as $\eqref{eqn:ratio}$ with shared parameter $\theta$:
\begin{equation}
\rt(\theta) = \frac{\pi_{\theta}\big(a_t^i|o_t^i\big)}{\pi_{\theta_{old}}\big(a_t^{i}|o_t^i\big)}.  
\end{equation}
Note that the PS technique naturally fits into the purpose of training a single policy to handle a variable number of agents, as the shared policy offers different actions for different agents based on observation from each agent.

Although a shared policy is suitable for a variable number of agents, having a variable number of agents leads to a variable length input to the NN. Thus, our policy network embed with an LSTM, a popular recurrent NN \cite{10.1162/neco.1997.9.8.1735}, which accepts a variable length input and produces a fixed-length output. 

Besides being able to take input vectors of variable length, another advantage of the LSTM for multi-agent coverage control is that it takes an \textit{ordered} sequence as input.
In our problem definition, the variable length part of the input is $\tilde{o}^i$, an agent's relative observation, which can be naturally treated as a sequence ordered according to the relative importance of other agents. 
In our application of the LSTM architecture, for every time step $t$, the first entry of $\tilde{o}^i_t$ will be fed first and produce a hidden state inside the LSTM. When the next entry is fed in, the LSTM will combine both the current entry and the hidden vector from the last entry to produce the next hidden vector. 
Eventually, the entire input $\tilde{o}^i$ is mapped to the final hidden vector, denoted as $h_t^i$, which represents an encoded vector that stores all the important information in $\tilde{o}^i$. 
Later entries have a more direct influence on the LSTM output. 

Therefore, we arrange the entries in $\tilde{o}^i$ in descending order of the norm. 
Such an ordering puts the relative position of the farthest agent first, and of the nearest agent last, and captures the intuition that nearby agents should have a stronger effect on the current agent's action. 
This ordering is also consistent with the agent interaction potential $U_I$, where two agents interact more strongly if they are closer to each other. 

The LSTM output $h^i_t$ is concatenated with the agents' internal state $s^i$ and fed into the standard MLP to output the policy distribution. The LSTM-based policy network is shown as the "Actor" part of Fig. \ref{fig:img1}.

\subsection{Self-Attention Based Value Decomposition Network}
\label{subsect:Critic}

With the CTDE framework, the shared critic network takes the global state $s$ as input to estimate the true state value, where $s$ is of variable length according to the number of agents. While one can attempt a similar procedure as in Sec. \ref{subsect:Actor} for the environment state and treat $s$ as a sequence of the agents' internal states $s^i$, the LSTM structure is not the most appropriate for the \textit{centralized} critic network since there is no natural ordering for agents. We propose instead a self-attention based value decomposition (VD) structure, illustrated in the ``Critic" part of Fig. \ref{fig:img1}, which is able to estimate the state value by equally assessing all agents' internal states $s^i$.

The self-attention was originally introduced in the Transformer architecture\cite{https://doi.org/10.48550/arxiv.1706.03762}, which is used to compute a sequence representation that associates the element at different positions in a sequence. For input a length-$n$ sequence $X\in \mathbb{R}^{n \times d_m}$ of vectors $x_i\in \mathbb{R}^{d_m}$, the self-attention of $X$ outputs
\begin{equation}
\label{eqn:attention}
\textit{\textbf{attention}}(X) = softmax\bigg(\frac{(X W_q) (X W_k)^\top}{\sqrt{d_k}}\bigg) (X W_v),
\end{equation}
\noindent where $W_q\in \mathbb{R}^{d_m \times d_k}, W_k\in \mathbb{R}^{d_m \times d_k}$, and $W_v \in \mathbb{R}^{d_m \times d_v}$ are learnable matrices that project every $x_i \in X$ from $\mathbb R^{d_m}$ onto $\mathbb R^{d_k}$, $\mathbb R^{d_k}$ and $\mathbb R^{d_v}$ respectively, and $\sqrt{d_k}$ is a scaling factor depending on the subspace dimension. Each row in the output of $\text{attention}(X) \in \mathbb{R}^{n \times d_v}$ can be viewed as a new vector representation for the $i$-th element $x_i$ in the sequence $X$, which contains the correlated information to another element in a different position.\\
\indent In our work, we use self-attention to encode the global state into a new representation for each agent and connect with a VD structure constructed by shared MLP. This idea is visualized in the "Critic" part of Fig. \ref{fig:img1}, where we treat the environment full state $s$ as a sequence of length $n$. For the critic input, the state $s_t$ at every time step is first sent into a linear layer that maps the internal state representation for each agent onto a higher dimensional space. Then, the output from the linear layer, of dimension $\mathbb{R}^{n \times d_m}$, is used to compute the attention output using the matrix equation \eqref{eqn:attention}. The final output from attention, denoted as $z_t = \textit{\textbf{attention}}(s_t)$, is considered a new representation of the environment. In particular, each row of $z_t$, denoted as $z^i_t$, is a new representation of $s^i_t$ that contains not only information of the intrinsic state for the $i$th agent but also the information about other agents relevant to coordination in the coverage task. In practice, the attention is often followed by residual connections and layer normalization to facilitate the training, and our implementation preserves this structure as well. 
Since the attention output fuses all agents' information while giving each agent different internal state representations, the shared MLP network can output different values according to the new representation $z^i$ for different agents. Such structure fits well with the VD approach \cite{VDN}, where we explicitly formulate the state value function estimation at time step $t$ as the summation of local state values estimated from every $z^i_t$, i.e.,
\begin{equation}
V_{\pi_{\theta}}(s_t) \approx \sum_{i=i}^{n} V_{\phi}(z^i_t) = \sum_{i=i}^{n} V_{\phi}\big(\textit{\textbf{attention}}(s_t)[\, i  \,]\big).
\end{equation}
We remark that the additive VD is legitimate in our setup due to the explicit additive reward formulation and the true value function defined as  
\begin{align*}
V_{\pi_{\theta}}(s_t) &= \mathbb{E}_{\pi_{\theta}}\big[G_t|s_t\big]\\
&= \mathbb{E}_{\pi_{\theta}} \bigg[\sum_{k=0}^{T} \sum_{i=1}^{n}\gamma^k R^i_{t+k} \bigg| s_t \bigg]\\
&= \sum_{i=1}^{n} \mathbb{E}_{\pi_{\theta}} \bigg[\sum_{k=0}^{T} \gamma^k R^i_{t+k} \bigg| s_t \bigg] = \sum_{i=i}^{n} V_{\pi_{\theta}}^i(s_t),
\end{align*}where $V_{\pi_{\theta}}^i(s_t)$ is the agents’ individual state value function and similar for the action value function defined as $Q_{\pi_{\theta}}(s_t, a_t) = \sum_{i=i}^{n} Q_{\pi_{\theta}}^i(s_t, a_t)$. The above setup yields the critic loss to be:
\begin{equation}
\label{eqn:cli}
L(\phi)  = \frac{1}{T}\sum_{t = 0}^{T}  \Big(\hat{V}_t^i - V_{\phi}(z^i_t) \Big)^2, \forall \; i \in \{1, \dots, n \},   
\end{equation}where the value target $\hat{V}_t^i$ is chosen to be the Monte-Carlo estimation of agents' individual returns. 
Note that the loss \eqref{eqn:cli} is different from the normal VD method, where the shared critic network directly estimates the true agent's individual state value instead of the global state value since the agent's individual rewards can be directly accessed in our work.

Moreover, VD helps to alleviate the credit assignment problem compared to using the global value function. VD allows us to decompose the advantage values $\hat{A}_t$ into the summation form as:
\begin{align*}
\hat{A}_t &= Q_{\pi_{\theta}}(s_t, a_t) - V_{\pi_{\theta}}(s_t)\\
&= \sum_{i=1}^{n} \Bigr[Q_{\pi_{\theta}}^{i}(s_t, a_t) - V_{\pi_{\theta}}^{i}(s_t)\Bigl]\\
&= \sum_{i=1}^{n} \hat{A}_t^i,
\end{align*}
where $\hat{A}_t^i$ denotes the agents' individual advantages over the current policy. The use of $\hat{A}_t^i$ gives each agent a better sense of their own contribution, which accelerates policy learning further. The structure of the shared value network is still an MLP, where the $z_t$ is fed to compute the final local state value for each agent. Note that the new structure of the critic network still agrees with the main feature of the CTDE framework, where a centralized critic network estimates the state value via global information.



\subsection{Training the Policy}
\label{subsect:TTP}
To smooth the policy training, we apply the strategy of two-phase training. The first phase is the pre-training step, and both the policy and critic networks could be fitted into the pre-training phase. For the policy network, the pre-training is applied using the BC method with a set of trajectories sampled from an expert policy $f$. Meanwhile, the critic network could be pre-trained with sampled trajectories using MSE loss with Monte-Carlo estimation of agents' individual returns as well.

After the pre-training step, the training shall be continued with the purely MAPPO algorithm based on the pre-trained initial model. In order to train a policy that can adapt to variable numbers of agents, the second phase requires sampling multiple environments across different numbers of agents, and this is done by paralleled sampling across multiple environments with different numbers of agents in each training iteration. Furthermore, for the possibility of transferring the policy into various domains, each paralleled environment is encouraged to use a different domain to ensure the training covers a variety of shapes.

\begin{algorithm}
    \caption{BC + Self-Attention Based MAPPO}
    \SetAlgoLined
    \SetKwInOut{Input}{input}
    \SetKwInOut{Output}{output}
    \DontPrintSemicolon
    \Input{A parameterized policy $\pi_{\theta}$ and self-attention based critic $V_{\phi}$.\linebreak
           A trajectory set $D_{\mathcal{E}} = \{(o^i_t, \hat{a}^i_t)\}$, sampled from an expert policy $f$.}
    
    
    Pre-training parameter $\theta_0$ by minimizing BC loss on trajectory set $D_{\mathcal{E}}$:\newline \hspace*{1.45em} $L^{BC}(\theta) = \frac{1}{|D_{\mathcal{E}}|}\sum_{(o^i_t, \hat{a}^i_t) \in D_{\mathcal{E}}} \bigr\|\hat{a}^i_t - \mu_{\pi_{\theta}}(o^i_t)\bigr\|^2.$\\[1pt]

    \For {$k = 0, 1, 2, \dots$} {
        Sample a set of trajectories $D_{\mathcal{T}} = \{(s_t, a_t)\}$ from current policy $\pi_{\theta_k}$.\\[2pt]
        
        Compute current individual state value estimation $V_{\phi_k}(z_t^i) = V_{\phi_k}\big(\textit{\textbf{attention}}(s_t)[\, i  \,] \big), \, \forall s_t \in D_{\mathcal{T}}$.\\[2pt]
        
        Estimate individual advantages $\hat{A}_t^i$ for current policy based on $V_{\phi_k}(z_t^i)$.\\[2pt]

        Compute the policy clip loss: \newline $J(\theta_k)  = \mathbb{E}_{\pi_{\theta_k}}\bigr[\Min (\frac{\pi_{\theta}(a_t^i|o_t^i)}{\pi_{\theta_k}(a_t^i|o_t^i)}\hat{A}_t^i, \newline \hspace*{6em} \; \text{clip}(\frac{\pi_{\theta}(a_t^i|o_t^i)}{\pi_{\theta_k}(a_t^i|o_t^i)}, 1-\epsilon, 1+\epsilon)\hat{A}_t^i \bigr]$.\\[2pt]
        
        Compute the critic loss: \newline $L(\phi_k)  = \frac{1}{n|D_{\mathcal{T}}|} \sum_{i=1}^{n} \sum_{t \in D_{\mathcal{T}}} \big(\hat{V}_t^i - V_{\phi_k}(z^i_t) \big)^2$.\\[2pt]
        
        Update policy parameter by gradient ascent: $\theta_{k+1} \gets \theta_{k} + \alpha \nabla_{\theta_k}J(\theta_k)$.\\[2pt]
        
        Update critic parameter by gradient decent: $\phi_{k+1} \gets \phi_{k} - \beta \nabla_{\phi_k}L(\phi_k)$.
    
    }
    \label{alg:PoEG}
\end{algorithm}
Since our reward function is partially sparse, most of the exploration in the early stage of training would only provide unimportant samples, as sparse rewards do not provide a meaningful signal to agents for most states. Using BC for model pre-training allows our policy to be trained toward expert policy at first, avoiding instability in the early training stage, while using the MAPPO algorithm to continue training upon the pre-trained initial model could leverage the main strengths of RL to obtain a potential better policy through proper exploration.

Moreover,  the expert policy chosen in this work is the control law proposed in \cite{CHACON202015167,https://doi.org/10.48550/arxiv.2009.12211}, which can be directly derived from the two potentials $U_h$ and $U_I$ (see \eqref{eqn:Vh} and \eqref{eqn:VI}) by taking the negative gradient with respect to each agent:
\begin{equation}
\label{eqn:fi}
a^i = - \nabla_i U_h(p^i) - \sum_{i=1}^{n} \nabla_i U_I(p^{i,j}) - cv^i,
\end{equation}
where $c$ is a positive constant. We remark that this controller drives system \eqref{eqn:dynamics} into an equilibrium configuration which is a critical point of the total system potential energy $\Phi$ from \eqref{eqn:energy-total}; note that critical points of $\Phi$ are $\rd$-\textit{subcover} configurations. For more discussion on the asymptotic behavior of system \eqref{eqn:dynamics} with this control law, we refer to \cite{CHACON202015167,https://doi.org/10.48550/arxiv.2009.12211}.


\section{SIMULATED EXPERIMENTS}
\label{sect:experiments}

In this section, we present simulation results of the policy obtained by our approach in various domains and across different numbers of agents. We will compare the performance of our policy and the state-of-the-art classical controller \eqref{eqn:fi} and show the remarkable adaptability of our policy over a different number of agents. 
 
Per our discussion in Sec. \ref{subsect:TTP}, the training process is split into two stages. In the pre-training stage, an initial policy is obtained using 10 thousand trajectories sampled from the classical controller \eqref{eqn:fi} over three particular agent counts, 6, 8, and 9 agents, with the square domain. The RL stage continues the training on the initial policy in multiple environments, using the same number of agent configurations but more types of domains as in the first phase. The domain in each environment is either a pre-generated random polygon or an equilateral polygon. We run 12 parallel environments across 6, 8, and 9 agents, each using a different domain shape. In every RL training iteration, we sample 4 trajectories from each environment, and the policy takes about 500 iterations to converge.

All simulations presented in this section are 30 seconds long. For all environments, the agent group's initial position is randomized but follows an approximately horizontal line configuration outside the domain (with only a slight offset from a perfect horizontal line) and with zero initial velocity.

\subsection{Time Efficiency over State-of-the-Art Classical Controller}
\begin{figure}[htbpt]
\vspace*{0.175cm}
\centerline{\includegraphics[width=0.45\textwidth]{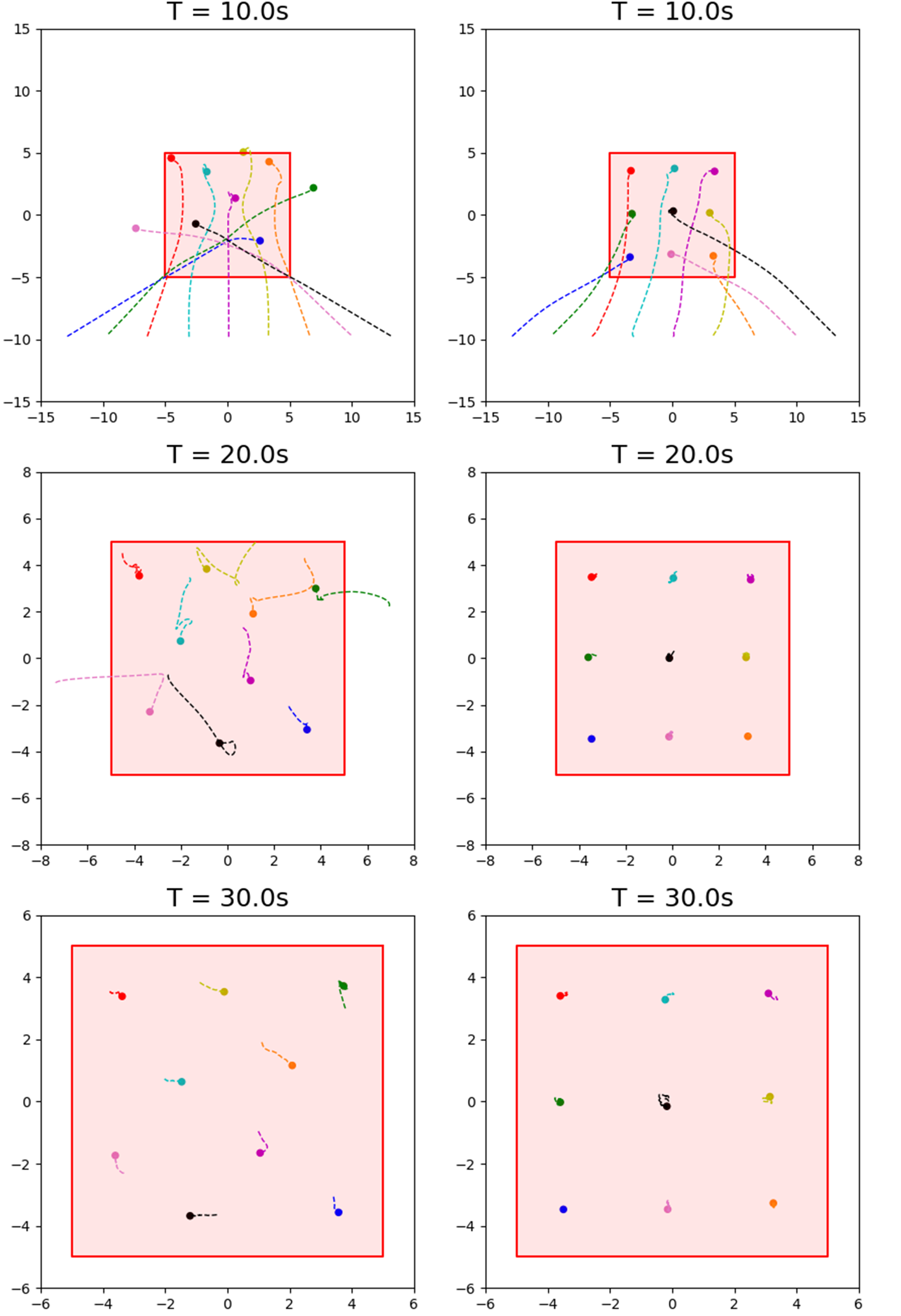}}
\caption{Square domain coverage of 9 agents for system \eqref{eqn:dynamics} with classical controller \eqref{eqn:fi} (left) and for our policy (right). 
Note the superior time efficiency of our policy, by which the agents converge to the correct $\rd$-\textit{subcover} configuration in only about 10 seconds.}
\label{fig:square}
\end{figure}

We compare our policy with the state-of-the-art controller in the 9-agent environment.  Take the simple square domain and note that the $\rd$-\textit{subcover} configuration for the square domain with 9 agents is simple and unique, in which case all agents are arranged in a square formation. Fig. \ref{fig:square} shows the agents' positions and trajectories generated by the classical controller \eqref{eqn:fi} (left column) and by our policy (right column). The colored dots represent the agents' positions at the current time, and the dashed tails represent the trajectories for the past 10 seconds. 

In the first 10 seconds, our policy behaves very differently from the controller \eqref{eqn:fi}. For the trajectories generated by the classical controller, an overshooting behavior can be observed. All agents initially enter the domain, but some pass through the domain after entering it. Such overshooting behavior reduces the time efficiency for achieving the final coverage configuration. In contrast, the trajectories from our policy show that all agents directly approach the final $\rd$-\textit{subcover} configuration. 
For the next 20 seconds, the classical controller brings the overshooting agents back into the domain, and then drives them slowly to the final configuration. Meanwhile, our policy maintains the $\rd$-\textit{subcover} configuration with only small perturbations over time.

\begin{figure}[htbp]
\centering
\vspace*{0.175cm}
\centerline{\includegraphics[width=0.475\textwidth]{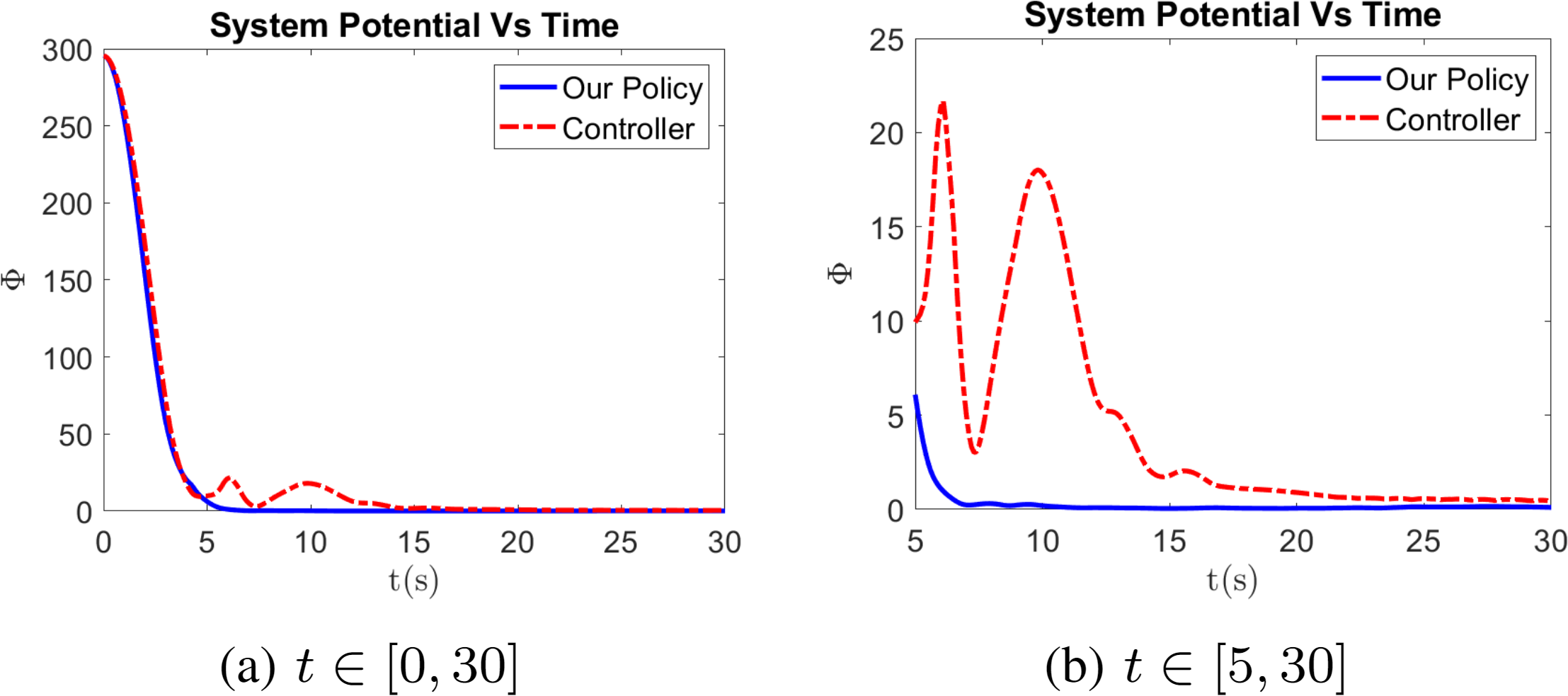}}
\caption{Time evolution of the total potential energy $\Phi$ from \eqref{eqn:energy-total} for the classical controller (dashed red) and our policy (solid blue). The left plot shows the entire duration, and the right plot zooms in on the last 25 seconds. The two bumps of the red curve in the left plot are due to strong repulsive forces and the overshooting behavior present in the classical controller, respectively. The right plot shows that our policy maintains a low potential in the last 20 seconds.  Note that the corresponding potential level of our policy is always below that of the classical controller.}
\label{fig:Phi}
\end{figure}

We also measure the coverage performance using the system potential $\Phi$. Fig. \ref{fig:Phi} shows the evolution of $\Phi$ over time, for the square domain simulation, where the red dashed curve corresponds to the classical controller \eqref{eqn:fi} and the solid blue curve is for our policy. The left plot is for the whole simulation duration, while the right plot gives a zoomed-in view of the last 20 seconds. During approximately the first 5 seconds, the system potential $\Phi$ decreases rapidly in both the classical controller \eqref{eqn:fi} and our policy. 

The two bumps of the potential energy for the classical controller (at around $t \approx 6$ 
and $t \approx 10$) can be explained as follows.  The first bump occurs when some agents get too close to each other inside the domain, resulting in strong repulsion forces. Such repulsive forces push the agents to get far away from each other so strongly that some agents are pushed out of the domain. This overshoot causes the second bump in $\Phi$ -- see also Fig. \ref{fig:square} (top left).

In contrast, the solid blue curve shows that for our policy, the potential decreases quickly and stays at a low level without any increase. Note that the potential level of our policy is always below that of the classical controller \eqref{eqn:fi}, which does not converge to the $\rd$-\textit{subcover} configuration within 30s duration.

\begin{figure}[ht]
\centerline{\includegraphics[width=0.45\textwidth]{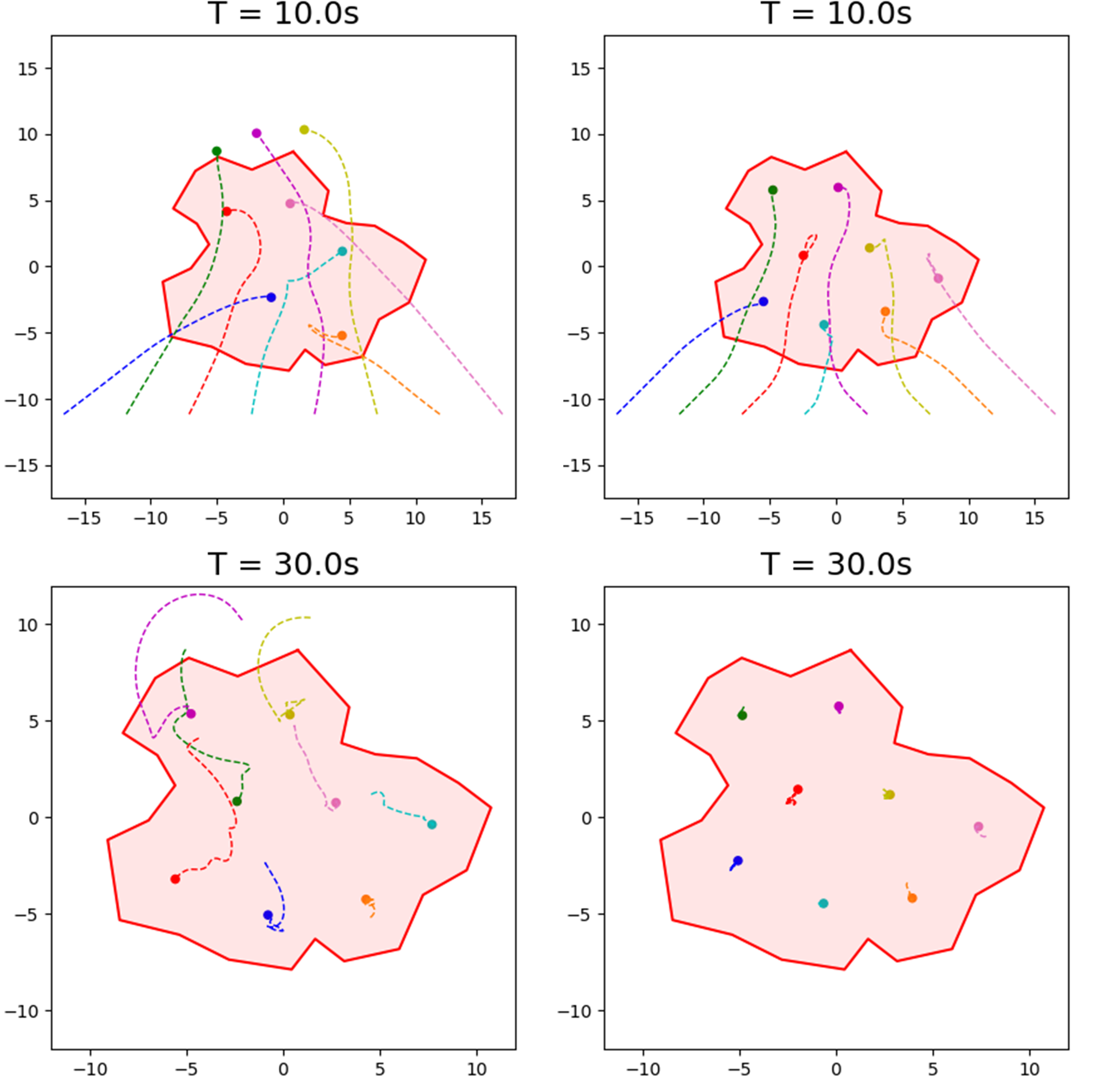}}
\caption{Non-convex domain coverage of 8 agents for system \eqref{eqn:dynamics} with classical controller \eqref{eqn:fi} (left) and for our policy (right). Our policy has a superior time efficiency for convergence to an $\rd$-\textit{subcover} configuration}
\label{fig:poly_8_agents}
\end{figure}

\begin{figure}[ht]
\centerline{\includegraphics[width=0.45\textwidth]{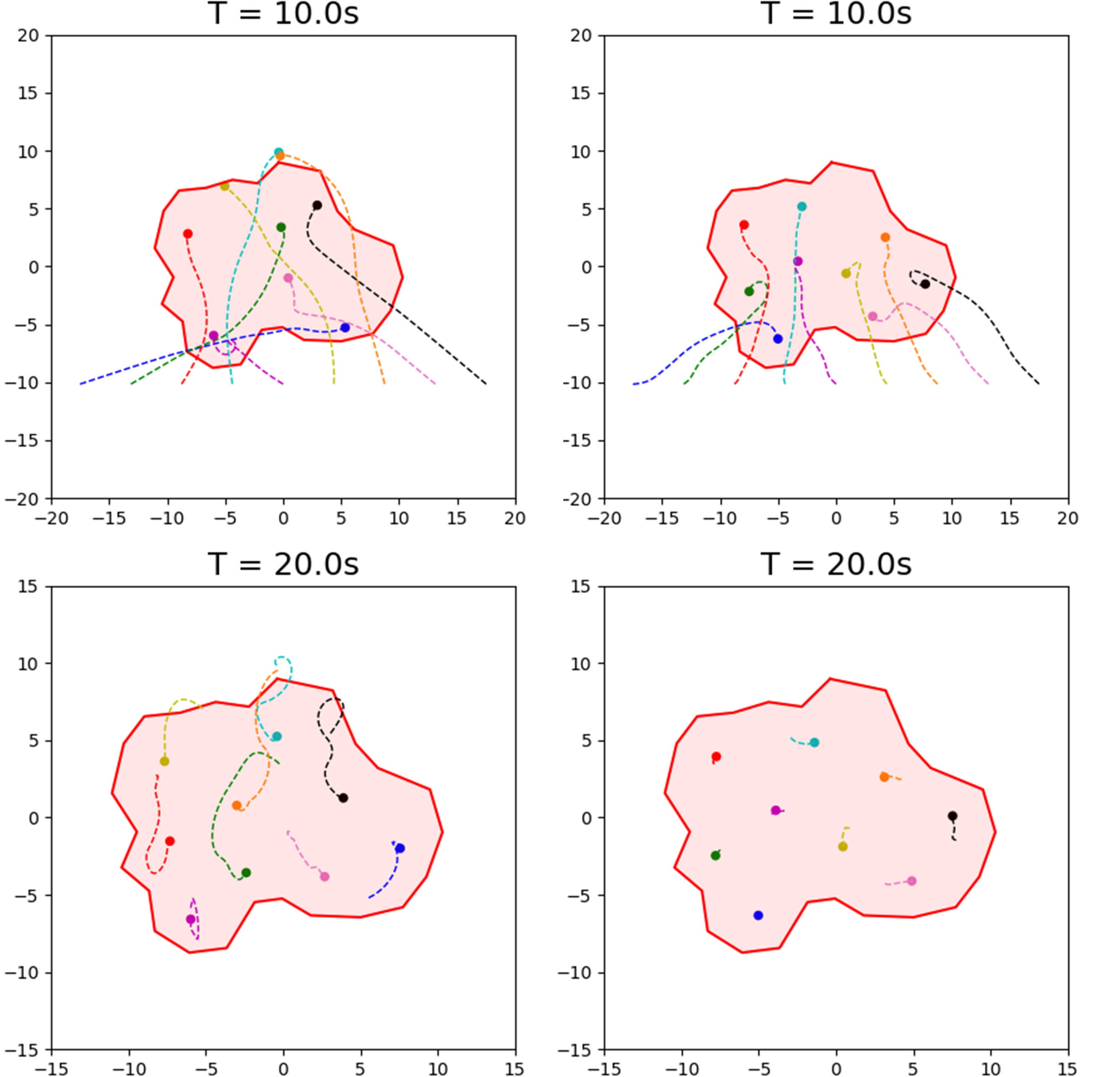}}
\caption{Non-convex domain coverage of 9 agents for system \eqref{eqn:dynamics} with classical controller \eqref{eqn:fi} (left) and for our policy (right). Our policy has a superior time efficiency for convergence to an $\rd$-\textit{subcover} configuration}
\label{fig:poly_9_agents}
\end{figure}

For non-convex non-symmetric random polygon domains, our policy preserves the same time-efficient advantages over the controller \eqref{eqn:fi}. Fig. \ref{fig:poly_8_agents} and Fig. \ref{fig:poly_9_agents} show the positions and trajectories of 8 and 9 agents that cover a non-convex, non-symmetric random polygon corresponding to the classical controller (left) and to our policy (right). Both the controller \eqref{eqn:fi} and our policy achieve the final coverage configuration within 20 seconds, while our policy only takes about 10 seconds, gaining a 35\% time efficiency improvement. We also note the overshooting phenomenon in the controller's behaviour for the non-convex polygon domain. 

Fig. \ref{fig:DifferentDomain} given more coverage trajectory for the last 15 seconds of the simulation time, including both convex and non-convex polygons, indicating the reliable coverage performance of our policy for both convex and non-convex cases.

\begin{figure}[htbp]
\centerline{\includegraphics[width=0.45\textwidth]{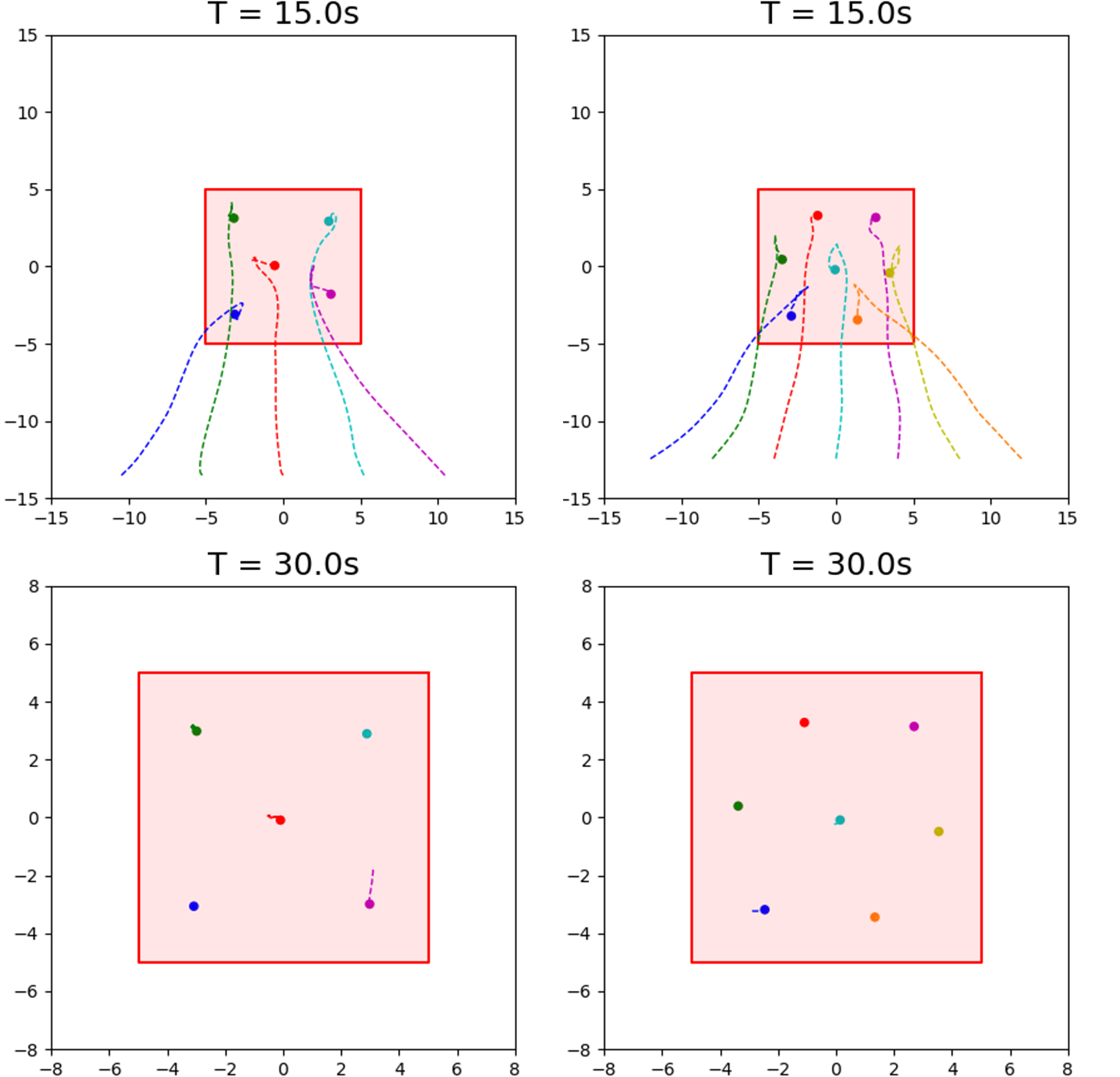}}
\caption{Zero-shot result of 5 and 7 agents on a square domain. The policy is directly deployed with unseen agent configurations and retains a satisfactory coverage result.}
\label{fig:zero_shot}
\end{figure}

\subsection{Generalization to Different Agent Counts}
The following experiment tests the policy adaptability from 5 to 9 agents. We remark that the cases of 5 and 7 agents do not appear during the training. These test cases aim to show the zero-shot generalizability of our approach.

We run 1000 simulations for each agent count (5 to 9) with both the classical controller \eqref{eqn:fi} and our policy in different domains (the domains are the same as for the RL training stage), and measure the system potential $\Phi$ at every time step for all sampled trajectories. A trajectory is considered to deliver a successful coverage result if the system potential level is below 0.15 (low enough to be an approximate $\rd$-\textit{subcover} configuration) before 30 seconds and lasts for the rest of the simulation time. 

Fig. \ref{fig:zero_shot} shows the policy behavior of 5 and 7 agents in the square domain. In both cases, our policy could attain the coverage configuration of about 15s and retain coverage for the rest of the simulation, which indicates the adaptability of the policy for unseen data. Table \ref{tab:1} shows the success rate of different agent counts over all simulations and the average convergence time over all samples that provided a successful coverage result. Our policy offers a compatible success rate with controller across different agent counts, with significant savings in the convergence times. Animated simulations are available on YouTube at \url{https://youtu.be/OgF62-NSbc0}.


\setlength{\tabcolsep}{0.8em}
\begin{table}[ht]
\begin{threeparttable}
\begin{tabular}{c cccc}
\toprule
Agents &\multicolumn{2}{c}{Success Rate}
&
\multicolumn{2}{c}{Convergence Time} \\\cmidrule(r){2-3}\cmidrule(l){4-5} & Controller& Policy    & Controller& Policy  \\
\midrule
  5 & 93.2\% &  85.3\% & 25.14s & 16.57s\\
  6 & 83.4\% &  88.6\% & 25.87s & 13.35s\\
  7 & 88.1\% &  82.7\% & 26.14s & 18.36s\\
  8 & 90.8\% &  94.1\% & 28.62s & 16.76s\\
  9 & 80.6\% &  83.6\% & 29.72s & 23.99s\\
\bottomrule
\end{tabular}
\caption{Success rate and average convergence time for controller \eqref{eqn:fi} and our policy. The success rate only considers the simulations that achieve a successful coverage result within 30 seconds. A "failed" simulation can possibly achieve the desired coverage result with a longer simulation time. Among all successful samples, our policy offers at least 20$\%$ improvements in the convergence speed for all agent counts.}
\label{tab:1}
\end{threeparttable}
\end{table}


\section{CONCLUSION}
In this work, we present a Self-Attention based reinforcement learning approach for the multi-agent efficient coverage control involving agents in continuous state and action with second-order dynamics. By using the LSTM and self-attention structure, our policy can perform adaptive control tasks with a variable number of agents. Our novel approach is also shown to outperform the existing classical controller in terms of time efficiency while maintaining the same level of coverage performance compared to the classical controller. Future work will investigate the potential of using reinforcement learning to solve coverage control incorporated with collision avoidance. 



\bibliography{bibliography.bib}

\begin{thebibliography}{10}

\bibitem{1}
G.~Zhang, G.~K. Fricke, and D.~P. Garg, ``Spill detection and perimeter surveillance via distributed swarming agents,'' {\em IEEE/ASME Trans. Mechatronics}, vol.~18, no.~1, pp.~121--129, 2013.

\bibitem{2}
W.~J. Yun, S.~Park, J.~Kim, M.~Shin, S.~Jung, D.~A. Mohaisen, and J.-H. Kim, ``Cooperative multiagent deep reinforcement learning for reliable surveillance via autonomous multi-uav control,'' {\em IEEE Trans. Industr. Inform.}, vol.~18, no.~10, pp.~7086--7096, 2022.

\bibitem{3}
X.~Dang, C.~Shao, and Z.~Hao, ``Target detection coverage algorithm based on 3d-voronoi partition for three-dimensional wireless sensor networks,'' {\em Mobile Information Systems}, vol.~2019, pp.~1--15, 03 2019.

\bibitem{4}
W.~Wang, V.~Srinivasan, K.-C. Chua, and B.~Wang, ``Energy-efficient coverage for target detection in wireless sensor networks,'' in {\em Int. Symposium on Information Processing in Sensor Networks}, 2007.

\bibitem{5}
D.~Drew, ``Multi-agent systems for search and rescue applications,'' {\em Current Robotics Reports}, vol.~2, June 2021.

\bibitem{6}
J.~P. Queralta, J.~Taipalmaa, B.~Can~Pullinen, V.~K. Sarker, T.~Nguyen~Gia, H.~Tenhunen, M.~Gabbouj, J.~Raitoharju, and T.~Westerlund, ``Collaborative multi-robot search and rescue: Planning, coordination, perception, and active vision,'' {\em IEEE Access}, vol.~8, pp.~191617--191643, 2020.

\bibitem{lo1}
J.~Cortes, S.~Martinez, T.~Karatas, and F.~Bullo, ``Coverage control for mobile sensing networks,'' {\em IEEE Trans. Robotics and Automation}, vol.~20, no.~2, pp.~243--255, 2004.

\bibitem{lo2}
Y.~Cao, W.~Yu, W.~Ren, and G.~Chen, ``An overview of recent progress in the study of distributed multi-agent coordination,'' {\em IEEE Trans. Industr. Inform.}, vol.~9, no.~1, pp.~427--438, 2013.

\bibitem{pfm1}
S.~S. Ge and Y.~Cui, ``Dynamic motion planning for mobile robots using potential field method,'' {\em Autonomous Robots}, vol.~13, pp.~207--222, 2002.

\bibitem{CHACON202015167}
J.~Chacon, M.~Chen, and R.~C. Fetecau, ``Safe coverage of compact domains for second order dynamical systems,'' {\em IFAC-PapersOnLine}, vol.~53, no.~2, pp.~15167--15173, 2020.

\bibitem{sfm2}
Y.~Tan, ``Multi-robot swarm for cooperative scalar field mapping,'' in {\em Handbook of Research on Design, Control, and Modeling of Swarm Robotics}, pp.~383--395, IGI Global, 2015.

\bibitem{sfm1}
M.~T. Nguyen, H.~M. La, and K.~A. Teague, ``Collaborative and compressed mobile sensing for data collection in distributed robotic networks,'' {\em IEEE Trans. Control Netw. Syst.}, vol.~5, no.~4, pp.~1729--1740, 2018.

\bibitem{MARL_COV_2}
H.~X. Pham, H.~M. La, D.~Feil-Seifer, and A.~Nefian, ``Cooperative and distributed reinforcement learning of drones for field coverage,'' 2018.
\newblock arXiv preprint:1803.07250.

\bibitem{MARL_COV_3}
J.~Xiao, G.~Wang, Y.~Zhang, and L.~Cheng, ``A distributed multi-agent dynamic area coverage algorithm based on reinforcement learning,'' {\em IEEE Access}, vol.~8, pp.~33511--33521, 2020.

\bibitem{MARL_COV_4}
J.~Heydari, O.~Saha, and V.~Ganapathy, ``Reinforcement learning-based coverage path planning with implicit cellular decomposition,'' 2021.
\newblock arXiv preprint: 2110.09018.

\bibitem{Adepegba2016MultiAgentAC}
A.~A. Adepegba, S.~Miah, and D.~Spinello, ``Multi-agent area coverage control using reinforcement learning,'' in {\em Int. Florida Artificial Intelligence Research Society Conf.}, 2016.

\bibitem{Kouzehgar_2020}
M.~Kouzehgar, M.~Meghjani, and R.~Bouffanais, ``Multi-agent reinforcement learning for dynamic ocean monitoring by a swarm of buoys,'' in {\em Global Oceans 2020: Singapore {\textendash} U.S. Gulf Coast}, 2020.

\bibitem{MADDPG}
R.~Lowe, Y.~Wu, A.~Tamar, J.~Harb, P.~Abbeel, and I.~Mordatch, ``Multi-agent actor-critic for mixed cooperative-competitive environments,'' in {\em Int. Conf. Neural Information Processing Systems}, December 2017.

\bibitem{https://doi.org/10.48550/arxiv.2009.12211}
J.~Chacon, M.~Chen, and R.~C. Fetecau, ``Safe coverage of moving domains for vehicles with second order dynamics,'' {\em IEEE Trans. Autom. Control}, 2022.
\newblock Early access.

\bibitem{10.5555/2967142}
F.~A. Oliehoek and C.~Amato, {\em A Concise Introduction to Decentralized POMDPs}.
\newblock Springer, 1st~ed., 2016.

\bibitem{MAPPO}
C.~Yu, A.~Velu, E.~Vinitsky, Y.~Wang, A.~Bayen, and Y.~Wu, ``The surprising effectiveness of {PPO} in cooperative, multi-agent games,'' 2022.
\newblock arXiv preprint: 2103.01955.

\bibitem{PPO}
J.~Schulman, F.~Wolski, P.~Dhariwal, A.~Radford, and O.~Klimov, ``Proximal policy optimization algorithms,'' 2017.
\newblock arXiv preprint: 1707.06347.

\bibitem{GAE}
J.~Schulman, P.~Moritz, S.~Levine, M.~Jordan, and P.~Abbeel, ``High-dimensional continuous control using generalized advantage estimation,'' in {\em Int. Conf. Learning Representations (ICLR)}, 2016.

\bibitem{10.1162/neco.1997.9.8.1735}
S.~Hochreiter and J.~Schmidhuber, ``{Long Short-Term Memory},'' {\em Neural Computation}, vol.~9, pp.~1735--1780, 11 1997.

\bibitem{https://doi.org/10.48550/arxiv.1706.03762}
A.~Vaswani, N.~Shazeer, N.~Parmar, J.~Uszkoreit, L.~Jones, A.~N. Gomez, L.~Kaiser, and I.~Polosukhin, ``Attention is all you need,'' in {\em Int. Conf. Neural Information Processing Systems}, 2017.

\bibitem{VDN}
P.~Sunehag, G.~Lever, A.~Gruslys, W.~M. Czarnecki, V.~Zambaldi, M.~Jaderberg, M.~Lanctot, N.~Sonnerat, J.~Z. Leibo, K.~Tuyls, and T.~Graepel, ``Value-decomposition networks for cooperative multi-agent learning based on team reward,'' in {\em Int. Conf. Autonomous Agents and MultiAgent Systems}, 2018.

\bibitem{10.1007/978-3-319-71682-4_5}
J.~K. Gupta, M.~Egorov, and M.~Kochenderfer, ``Cooperative multi-agent control using deep reinforcement learning,'' in {\em Autonomous Agents and Multiagent Systems}, 2017.

\bibitem{https://doi.org/10.48550/arxiv.2005.13625}
J.~K. Terry, N.~Grammel, S.~Son, and B.~Black, ``Parameter sharing for heterogeneous agents in multi-agent reinforcement learning,'' 2020.
\newblock arXiv preprint: 2005.13625.

\end{thebibliography}

\end{document}